\def\year{2020}\relax
\documentclass[letterpaper]{article} 
\usepackage{aaai20}  
\usepackage{times}  
\usepackage{helvet} 
\usepackage{courier}  
\usepackage[hyphens]{url}  
\usepackage{graphicx} 
\usepackage{graphicx}  
\usepackage{amsmath}
\usepackage{bm}
\usepackage{subfigure}
\usepackage{xspace}
\usepackage{multirow}
\usepackage{booktabs}

\newcommand{\RE}{\texttt{RE}\xspace}
\newcommand{\NRE}{\texttt{NRE}\xspace}
\newcommand{\KB}{\texttt{KB}\xspace}
\newcommand{\NN}{\texttt{NN}\xspace}
\newcommand{\NNs}{\texttt{NNs}\xspace}
\newcommand{\ILP}{\textit{ILP}\xspace}
\newcommand{\CLC}{\textit{CLC}\xspace}
\newcommand{\ACNN}{\textit{ACNN}\xspace}
\newcommand{\APCNN}{\textit{APCNN}\xspace}
\newcommand{\ConstraintLoss}{\textit{Constraint Loss}\xspace}
\newcommand{\Coherent}{\textit{Coherent}\xspace}
\newcommand{\Semantic}{\textit{Semantic}\xspace}
\newcommand{\citet}[1]{\citeauthor{#1} \shortcite{#1}}
\newcommand{\citep}{\cite}
\newcommand{\citealp}[1]{\citeauthor{#1} \citeyear{#1}}

\begin{document}
%
\setlength\titlebox{2.5in} 
\title{Integrating Relation Constraints with Neural Relation Extractors}
\author{Yuan Ye, Yansong Feng\textsuperscript{*}, Bingfeng Luo, \\
	{\bf \Large Yuxuan Lai, Dongyan Zhao} \\
	Wangxuan Institute of Computer Technology, Peking University, China\\
	\{pkuyeyuan, fengyansong, bf\_luo, erutan, zhaodongyan\}@pku.edu.cn\\}
\maketitle

\begin{abstract}
Recent years have seen rapid progress in identifying predefined relationship between entity pairs using neural networks (\NNs). 
However, such models often make predictions for each entity pair individually, 
thus often fail to solve the inconsistency among different predictions, 
which can be characterized by discrete relation constraints.
These constraints are often defined over combinations of entity-relation-entity triples, 
since there often lack of explicitly well-defined type and cardinality requirements for the relations.
In this paper, we propose a unified framework to integrate relation constraints with \NNs by introducing a new loss term, \ConstraintLoss. 
Particularly, we develop two efficient methods to capture how well the local predictions from multiple instance pairs satisfy the relation constraints.
Experiments on both English and Chinese datasets show that 
our approach can help \NNs learn from discrete relation constraints to reduce inconsistency among local predictions, 
and outperform popular neural relation extraction (\NRE) models even enhanced with extra post-processing. 
Our source code and datasets will be released at \url{https://github.com/PKUYeYuan/Constraint-Loss-AAAI-2020}.
\end{abstract}

\section{Introduction}
\label{introduction}
Relation extraction (\RE) aims to extract predefined relations between two marked entities in plain texts, 
and its success can benefit many knowledge base (\KB) related tasks 
like knowledge base population (\texttt{KBP})~\cite{suchanek2013advances,wu2018entity}, 
question answering (\texttt{QA})~\cite{dai2016cfo,yu2017acl,lai2019lattice} and etc.

Most existing works investigate the \RE task in a classification style. 
A sentence marked with a given pair of entities is fed to a classifier to decide their relationship, also called the \textit{sentence-level} \RE. 
Another related setup is to feed a group of sentences containing the given entity pair to the classifier, called the \textit{bag-level} \RE.  
We should note that both \textit{sentence-level} \RE and \textit{bag-level} \RE make predictions for each entity pair individually and locally. 
However, when we look at the model outputs globally, there are always contradictions among different predictions, 
such as an entity is regarded as the object of both \textit{Country} and \textit{City}, 
two different cities have been labeled as \textit{Capital} for one country and so on.
To alleviate these local contradictions, \citealp{chen2018encoding} collect constraints on the type and cardinality requirements of relations, 
such as whether two relations should not have the same type of subject (object), 
or whether a relation should not have multiple subjects (objects) given its object (subject). 
Further, in the inference stage, they use integer linear programming (\ILP) to filter and adjust the local predictions that are inconsistent with these constraints.
Basically, \ILP operates in a post-processing way to copy with contradictory predictions, but there is no way to provide feedback to the original \RE model. 

In fact, it would be of great importance to utilize those constraints to backwards improve the original \RE models. 
For example, enhanced with various attention or pooling mechanisms, 
most current neural network extraction models have shown promising performance on benchmark datasets, 
but they still suffer from inconsistent local predictions \cite{chen2018encoding}. 
If those relation constraints can be learned during model training, that will help to further improve  the overall performance, 
and we may no longer need a separate post-processing step as \ILP does. 

However, directly integrating relation constraints with \NRE models is not a trivial task: 
(1) relation constraints are not defined regarding a single prediction, but often  over combinations of instances, 
thus it is not easy to find appropriate representations for those constraints; 
(2) it is not easy to evaluate how well pairwise predictions match the constraints in a batch, 
and it is not clear how to feed the information back to the \NRE models.

To tackle the challenges, we propose a unified framework to flexibly integrate relation constraints with \NRE models by introducing a loss term \ConstraintLoss. 
Concretely, we develop two methods denoted as \Coherent and \Semantic to construct \ConstraintLoss from different perspectives.
\Coherent captures how well pairwise predictions match the constraints from an overall perspective, 
and \Semantic pays more attention to which specific rule in the constraints the pairwise predictions should satisfy.
In addition, we encode relation constraints into different representations for each method.
Notably, \ConstraintLoss is regarded as a rule-based regularization term within a batch instead of regularizing each instance, 
since the relation constraints are often defined over combinations of local predictions. 
Moreover, our approach does not bring extra cost to the inference phase and can be adapted to most existing \NRE models without explicit modifications to their structures,  
as it only utilizes the outputs from the \NRE model as well as relation constraints to obtain \ConstraintLoss 
and provides feedback to the \NRE model through backward propagation. 
Experiments on both Chinese and English datasets show that our approach can help popular \NRE models learn from the constraints and outperforms state-of-the-art methods even enhanced with \ILP post-processing. 
Moreover, jointing our approach and \ILP achieves further improvement which demonstrates that our approach and the \ILP post-processing exploit complementary aspects from the constraints.

The main contributions of this paper include: 
(1) We propose a unified framework to effectively integrate \NRE models with relation constraints without interfering the inherent \NRE structure. 
(2) We develop two efficient methods to capture  the inconsistency between local \NRE outputs and relation constraints, which are used as a loss term to help the \NRE training.   
(3) We provide thoroughly experimental study on different datasets and base models. The results show that our approach is effective and exploits the constraints from different perspectives with \ILP.

\section{Related Work}
\label{sec:related_work}
Since annotating high-quality relational facts in sentences is laborious and time-consuming, 
\RE is usually investigated in the distant supervision (\texttt{DS}) paradigm, 
where datasets are automatically constructed by aligning existing \KB triples $<subj, r, obj>$
\footnote{We use $subj$, $obj$ and $r$ to denote subject, object and relation for a \KB triple, respectively, in the rest of this paper.} 
with a large text corpus ~\cite{mintz2009distant}. 
However, the automatically constructed dataset suffers the wrong labeling problem, 
where the sentence that mentions the two target entities may not express the relation they hold in \KB, 
thus contains many false positive labels \cite{riedel2010modeling}. 
To alleviate the wrong labeling problem,
\RE is usually investigated in the multi-instance learning (\texttt{MIL}) framework which considers \RE task at bag-level and holds the at-least-one hypothesis, 
thinking that there exists at least one sentence which expresses the entity pair's relation in its corresponding sentence bag \cite{hoffmann2011knowledge,surdeanu2012multi,suchanek2013advances}.

As neural networks have been widely used, 
an increasing number of researches for \RE have been proposed under \texttt{MIL} framework. 
\citealp{zeng2014relation} use a convolution neural network (\textit{CNN}) to automatically extract features and \citealp{zeng2015distant} use a piece-wise convolution neural network (\textit{PCNN}) to capture structural information by inherently splitting a sentence into three segments according to the two target entities. 
Furthermore, \citealp{lin2016neural} proposed sentence-level attention-based models (\ACNN, \APCNN) to dynamically reduce the weights of noisy sentences.
And there also exists many \NN based works improving the \RE performance by utilizing external information, 
such as syntactic information \cite{he2018see}, entity description \cite{ji2017distant}, relation aliases \cite{vashishth2018reside} and etc.

In addition, there are many works focusing on combining \NNs with precise logic rules to harness flexibility and reduce uninterpretability of the neural models. 
\citealp{hu2016harnessing} make use of first-order logic (FOL) to express the constraints and propose a teacher-student network that could project prediction probability into a rule-regularized subspace and transfer the information of logic rules into the weights of neural models.
\citealp{DBLP:conf/icml/XuZFLB18} put forward a semantic loss framework, which bridges between neural output vectors and logical constraints by evaluating how close the neural network is to satisfying the constraints on its output with a loss term. 
And \citealp{luo2018marrying}  develop novel methods to exploit the rich expressiveness of regular expressions at different levels within a \NN, 
showing that the combination significantly enhances the learning effectiveness when a small number of training examples are available.

However, using these frameworks on \RE is not straightforward.
Specifically, \citealp{hu2016harnessing} directly project prediction probability of instance as they can assess how well a single instance's prediction satisfies the rules, 
while constraints in \RE are non-local and we could not examine each instance individually for the violation of constraints.
\citealp{luo2018marrying} need the regular expressions to provide keyword information and get a priori category prediction, 
however, generating high-quality regular expressions from \RE datasets is not easy.
For \citealp{DBLP:conf/icml/XuZFLB18}, since our constraints are related to the combination of instances rather than a single instance, 
to utilize the semantic loss framework, we need to find appropriate representations for various relation constraints and evaluate the neural output in a pairwise way.

\section{Relation Constraints}
\label{sec:constraints}
Since many \texttt{KBs} do not have a well-defined typing system and explicit argument cardinality requirements, 
similar in spirit with \citealp{chen2018encoding}, 
our relation constraints are defined over the combination of two triples:$<subj_m, r_m, obj_m>$ and $<subj_n, r_n, obj_n>$.
\footnote{The main difference is that our constraints are considered as positive rules where we expect the relation predictions to fall in, while the constraints in \citealp{chen2018encoding} are considered as inviolate rules that the local predictions should not break.}
We derive the type and cardinality constraints from existing \KB to implicitly capture the expected type and cardinality requirements on the arguments of a relation. One can surely employ human annotators to collect such constraints.

\paragraph{Type Constraints.}
Type constraints implicitly express the types of subjects and objects that a specific relation could have. 
For example, the subject and object types for relation \textit{almaMater} should be \textsc{person} and \textsc{school}, respectively,
and we take positive rules \textbf{[\textit{almaMater} and \textit{knownFor} could have the same subject type]} 
and \textbf{[\textit{almaMater} and \textit{employer} could have the same object type]} 
to implicitly encode \textit{almaMater}'s subject and object type requirements.

Specifically, we use entity sharing between different relations to implicitly capture the expected argument type of each relation.
If the subject (or object) set of relation $r_i$ in \KB has an intersection with those of $r_j$, 
then we consider $r_i$ and $r_j$ could have the same expected subject (or object) type.
We thereby assign relation pairs ($r_i$, $r_j$) into $\bm{C^{ts}}$ if they are expected to have the same subject type, 
into $\bm{C^{to}}$ if they are expected to have the same object type, 
and assign it into $\bm{C^{tso}}$  if the subject type of one relation is expected to be same as the object type of the other. 

\paragraph{Cardinality Constraints.}
Cardinality constraints indicate the cardinality requirements on a relation's arguments. 
For example, relation \textit{almaMater} could have multiple subjects (\textsc{person}) when its object (\textsc{school}) is given.

Specifically, for each predefined relation $r_i$, we collect all triples containing $r_i$, 
and count the number of the triples that have multiple objects (subjects) for each subject (object). 
Then, we assign relation $r_i$ into $\bm{C^{cs}}$ if it can have multiple subjects for a given object, 
into $\bm{C^{co}}$ if it can have multiple objects for a given subject.

Finally, we get 5 sub-category constraint sets.
We use $\bm{C^{\phi}}$ to represent a single set, 
$\bm{C^{t*}} \in \{\bm{C^{ts}, C^{to}, C^{tso}}\}$ to represent a  type constraint set, 
and $\bm{C^{c*}} \in \{\bm{C^{cs}, C^{co}}\}$ to represent a cardinality constraint set.
Note that  our relation constraints are defined to examine whether a pair of subject-relation-object triples can hold at the same time from different perspectives.
To make our constraints clearer, we list some rules for each constraint set in Table \ref{Table:example_rules}.

\begin{table}[htbp]
	\centering
	\small
	\begin{tabular}{ll}
		\toprule[1pt]
		\textbf{Set} & \textbf{Sampled Positive Rules}                                                                                                                                                                                                                             \\ \midrule[0.75pt]
		$\bm{C^{ts}}$  & \textit{\begin{tabular}[c]{@{}l@{}}(almaMater, knowFor), (city, region), (spouse, child) \end{tabular}}   \\
		$\bm{C^{to}}$  & \textit{\begin{tabular}[c]{@{}l@{}}(almaMater, owner), (city, hometown), (capital, city) \end{tabular}}   \\ 
		$\bm{C^{tso}}$ & \textit{\begin{tabular}[c]{@{}l@{}}(birthPlace, capital), (child, spouse), (city, country) \end{tabular}} \\
		$\bm{C^{cs}}$  & \textit{\begin{tabular}[c]{@{}l@{}}almaMater, country, city, hometown \end{tabular}} \\
		$\bm{C^{co}}$  & \textit{\begin{tabular}[c]{@{}l@{}}foundationPerson, child, knownFor, product \end{tabular}} \\ 
		\bottomrule[1pt]
	\end{tabular}
	\caption{Example rules for each constraint set $\bm{C^{\phi}}$.}
	\label{Table:example_rules}
\end{table}

\section{Our Approach}
\label{sec:approach}
As shown in Fig.~\ref{fig:framework}, our framework consists of two main components, 
a base \NRE model and the Constraint Loss Calculator (\CLC). The \CLC module is designed to integrate the relation constraints with \NRE models, 
which does not rely on specific \NRE architectures and can work in a plug-and-play fashion.

\subsection{Base NRE Model}
\label{sec:base_model}
While our framework can work with most existing relation extractors, in this paper, 
we take the most popular neural relation extractors, \ACNN and \APCNN ~\cite{lin2016neural}, as our base extractors.
\footnote{We do not use the most recently neural models, such as \citealp{feng2018reinforcement}, \citealp{qin2018dsgan} and \citealp{jia2019arnor}, as our base model, as they focused more on noise reduction which is not within the scope of this paper.}

\textbf{\ACNN} uses convolution neural networks with max-pooling layer to capture the most significant features from a sentence. 
Then, an attention layer is used to selectively aggregate individual representations from a bag of sentences into a sentence bag embedding, 
which is fed to a softmax classifier to predict the relation distribution $\bm{p_\theta(Y|X)}$.

\textbf{\APCNN} is an extension of \ACNN. 
Specifically, \APCNN divides the convolution output into three segments based on the positions of the two given entities and devises a piece-wise max-pooling layer to produce sentence representation.

\subsection{Constraint Loss Calculator (CLC)}
\label{sec:framework}

\begin{figure}
	\centering
	\includegraphics[width=7.5cm]{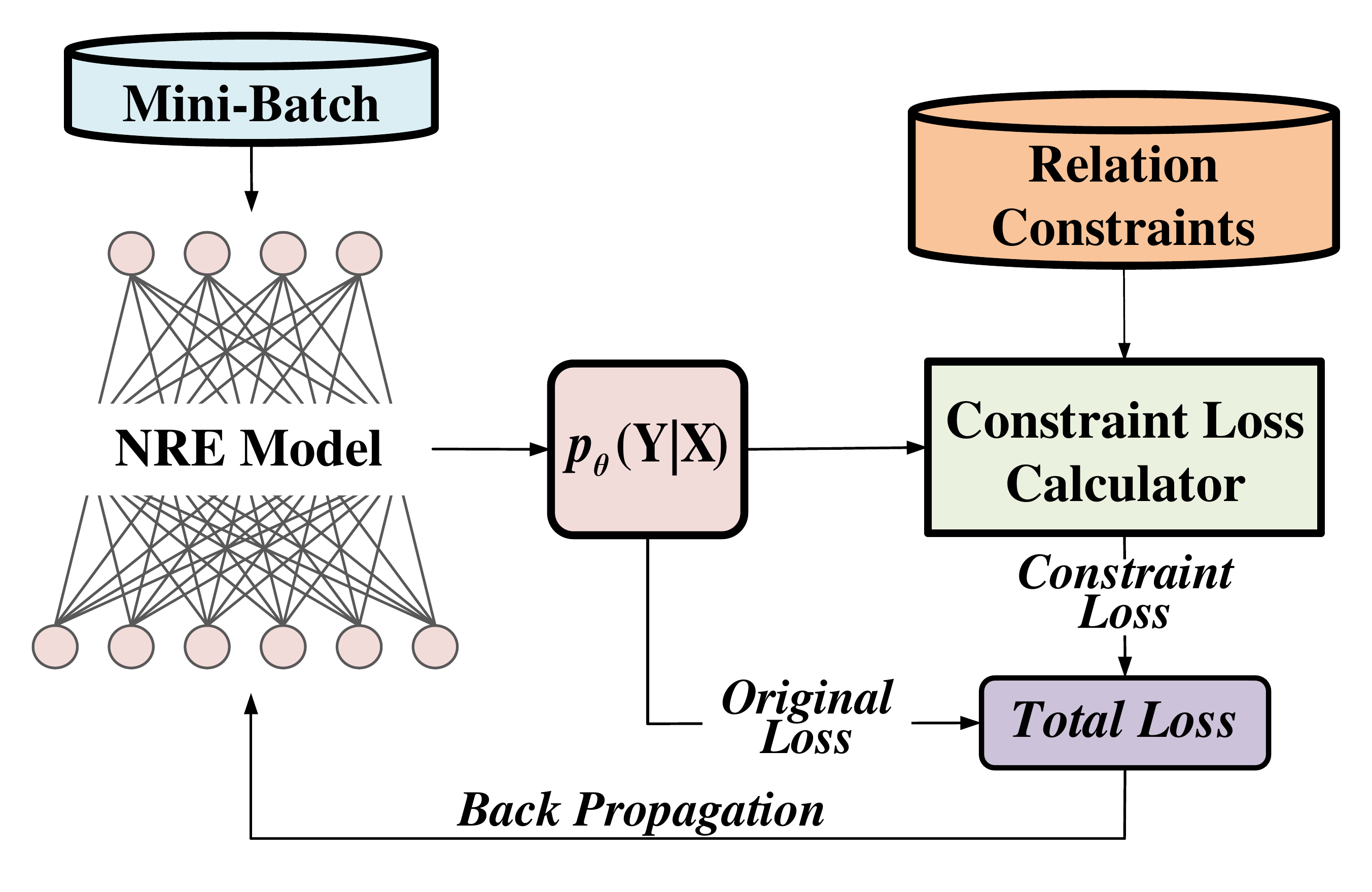}
	\caption{Framework overview. For each mini-batch, the \ConstraintLoss is calculated by evaluating the predicted probability $\bm{p_\theta(Y|X)}$ according to the relation constraints.}
	\label{fig:framework}
\end{figure}

\begin{figure*}
	\centering
	\includegraphics[width=16.0cm]{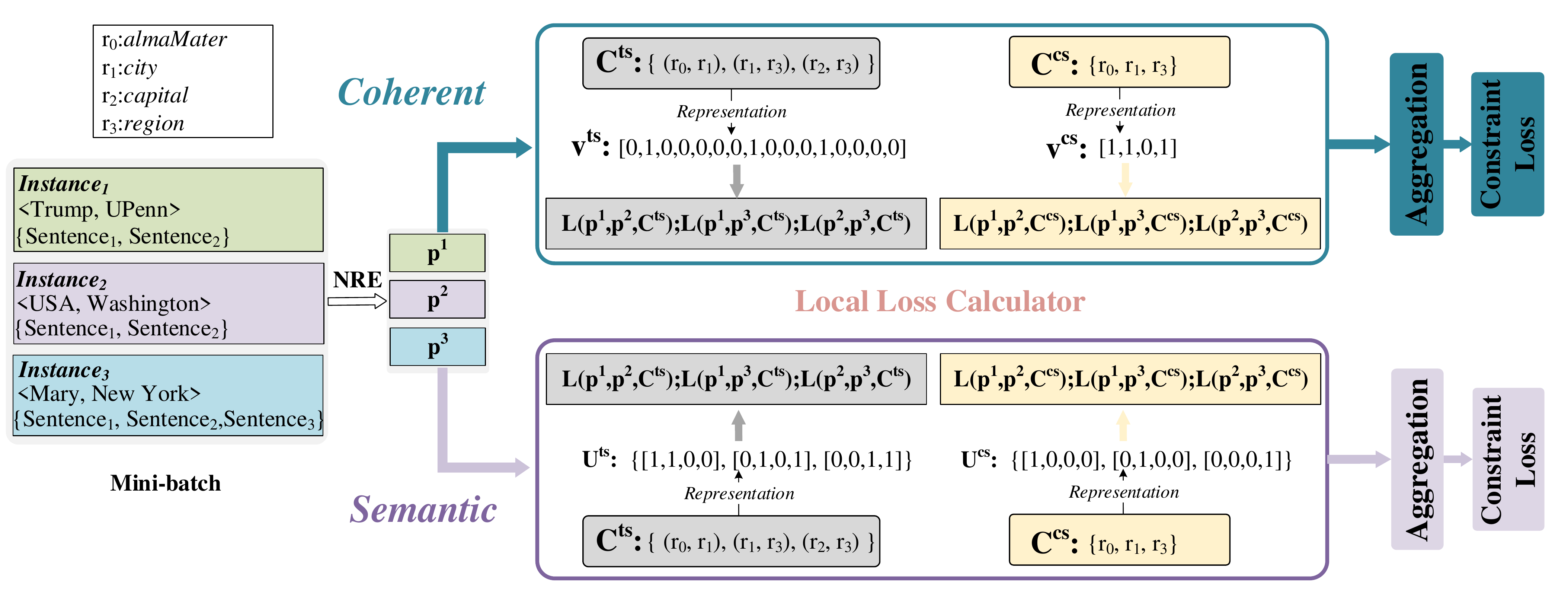}
	\caption{A running example of our two \CLC modules, \textit{Semancit} and \textit{Coherent}. 
			To exhibit the main process clearly, we simplify the example and only consider 4 relations, 
			a mini-batch with 3 instances and two constraint sets ($\bm{C^{ts}}$, $\bm{C^{cs}}$). 
			The whole process of \CLC contains 3 steps, we take \textit{Semantic} as an example. 
			First, we represent constraint set $\bm{C^{ts}}$ ($\bm{C^{cs}}$) as a vector set $\bm{U^{ts}}$ ($\bm{U^{cs}}$) and each vector represents a single rule. 
			Then, we feed \NRE output and the vector set into local loss calculator, 
			getting the local loss $\bm{L(p^m, p^n, C^{\phi})}$ (using Eq. \ref{eq:semantic_loss}) for each pair of instances within a batch. 
			Finally, Constraint Loss is obtained by aggregating all instance pairs in the batch. 
			The main difference between \textit{Semantic} and \textit{Coherent} is that \textit{Coherent} represents constraint set into \textbf{one vector} while \textit{Semantic} represents it into a \textbf{vector set}, utilizing relation constraints from different perspectives.}
	\label{fig:CLC_details}
\end{figure*}

Given the inherent nature of our relation constraints, 
the \CLC can not evaluate a single subject-relation-object prediction against our constraint sets, 
we thus operate our \CLC in a mini-batch wise way.
Specifically, in each batch, we integrate the relation constraints with the \NRE output by introducing a loss term, \ConstraintLoss, 
which is designed to regulate the \NRE model to learn from those constraints, \textbf{e.g., not to violate the positive rules in the constraints}.
As shown in Fig. \ref{fig:framework}, to calculate \ConstraintLoss, 
we first collect the \NRE output probability $\bm{p_\theta(Y|X)}$ within the batch, 
and then the \CLC takes $\bm{p_\theta(Y|X)}$ and relation constraints as input to obtain \ConstraintLoss,
which should reflect the inconsistency among all local predictions $\bm{p_\theta(Y|X)}$ according to our relation constraints.
Finally, the total loss for back propagation consists of two parts: the original \NRE loss ($L_O$) and the \ConstraintLoss ($L_C$): 
$$ L_{total} = L_{O} + \lambda L_{C} $$
where $\lambda$ is a weight coefficient.

Particularly, the key task of \CLC is to evaluate how well the current \NRE output probabilities $\bm{p_\theta(Y|X)}$ satisfy our relation constraints. 
We solve this problem in two steps.
We first calculate a local loss term $L(\bm{p^m}, \bm{p^n}, \bm{C^{\phi}})$ for a pair of local predictions, 
i.e., $\bm{p^m}$ and $\bm{p^n}$ for the $m^{th}$ and $n^{th}$ instances, respectively,
\footnote{We use $\bm{p^m}$ to represent the probability output of the neural model on the $m^{th}$ instance in a batch.}
against the constraint set $\bm{C^{\phi}}$.
Secondly, we aggregate all local loss terms to obtain the batch-wise \ConstraintLoss.
Here, we develop two methods to calculate \ConstraintLoss from different perspectives, 
denoted as \textit{Coherent (Coh)} and \textit{Semantic (Sem)}, respectively.

\subsection{\textit{Coherent (Coh)}}
In this method, we calculate \ConstraintLoss by  evaluating the coherence between the \NRE output  and a constraint set.
Note that this method only requires the \NRE outputs to be more consistent with one constraint set as a whole, 
but does not explicitly push the \NRE model to update according to specific positive rules in this set.

\paragraph{Representing Constraint Sets.} 
We encode a constraint set into \textbf{one} single binary vector. 
Since the positive rules in the type and cardinality constraint set have different forms, 
we represent them in slightly different ways.

For a type constraint set $\bm{C^{t*}}$, we construct a binary vector $\bm{v^{t*}}$, 
where $v^{t*}_{i,j}$ indicates whether relation pair $(r_i, r_j)$ belongs to $\bm{C^{t*}}$, 
i.e., $v^{t*}_{i,j}=1$  if $(r_i, r_j) \in \bm{C^{t*}}$ and $v^{t*}_{i,j}=0$ if $(r_i, r_j) \not \in \bm{C^{t*}}$.
Take $\bm{C^{ts}}$ illustrated in Fig.~\ref{fig:CLC_details} as an example,
since $(\textit{almaMater}, \textit{city}) \in \bm{C^{ts}}$, $v^{ts}_{0,1}$ is set to 1. 

For a cardinality constraint set $\bm{C^{c*}}$, we construct a binary vector $\bm{v^{c*}}$, 
where $v^{c*}_{i}$ indicates whether relation $r_i$ belongs to $\bm{C^{c*}}$, 
i.e., $v^{c*}_{i}=1$ if $r_i \in \bm{C^{c*}}$ and $v^{c*}_{i}=0$ if $r_i \not \in \bm{C^{c*}}$.
Again, in Fig.\ref{fig:CLC_details}, $v^{cs}_{0}$ is set to 1, since $\textit{almaMater} \in \bm{C^{cs}}$. 

Thus, for each one of the 5 sub-category constraint sets, we build one single representation vector, 
resulting in 5 vectors $\{\bm{v^{ts}},\bm{v^{to}}, \bm{v^{tso}}, \bm{v^{cs}}, \bm{v^{co}}\}$. 
And the dimensions of $\bm{v^{t*}}$ and $\bm{v^{c*}}$ are $|R|^2$ and $|R|$, respectively, 
where $|R|$ is the size of the relation set.

\paragraph{Local Loss Calculation.} 
Now, we proceed to calculate the loss term for a pair of local predictions, 
e.g., the $m^{th}$ and $n^{th}$ instances, within a batch.
Our expectation is that coherent local prediction pairs should  satisfy our constraint sets.
Again, we deal with the type constraint sets and cardinality constraint sets separately.

Thus, for a type constraint set $\bm{C^{t*}}$ represented by $\bm{v^{t*}}$, the local loss, $L(\bm{p^m},\bm{p^n},\bm{C^{t*}})$,
can be written as:
\begin{equation}
\label{eq:coh_type_calculation}
L(\bm{p^m}, \bm{p^n}, \bm{C^{t*}}) =  -I^{mn}_{t*}log(\sum_{i,j}v^{t*}_{i,j}p^m_i p^n_j)
\end{equation}
where $I^{mn}_{t*}\in\{0, 1\}$ indicates whether to calculate $L(\bm{p^m}, \bm{p^n}, \bm{C^{t*}})$. Take $I^{mn}_{ts}$ as an example, 
for triple pair $(subj_m, r_m, obj_m)$ and $(subj_n, r_n, obj_n)$, 
we set $I^{mn}_{ts}=1$, if $subj_m=subj_n$ which means the two triples have the same subject type and corresponding predicted relation pair should satisfy $\bm{C^{ts}}$; 
otherwise, we assign 0 to $I^{mn}_{ts}$.\footnote{Detailed assignment for $I^{mn}_{\phi}$ can be found in Appendix.}
$p^m_i p^n_j$ can be considered as the probability that the base \NRE model predicts relation $r_i$ and $r_j$ for the $m^{th}$ and $n^{th}$ instances, respectively.

For cardinality constraint set $\bm{C^{c*}}$ represented by $\bm{v^{c*}}$, the local loss, $L(\bm{p^m},\bm{p^n},\bm{C^{c*}})$, can be written as:
\begin{equation}
\label{eq:coh_cardinality_calculation}
L(\bm{p^m}, \bm{p^n}, \bm{C^{c*}}) =  -I^{mn}_{c*}log(\sum_{i}v^{c*}_ip^m_i p^n_i)
\end{equation}
where $I^{mn}_{c*}$ is an indicator similar to $I^{mn}_{t*}$ and $p^m_i p^n_i$ is seen as the possibility that the base \NRE predicts relation $r_i$ for both the $m^{th}$ and $n^{th}$ instances.

\paragraph{Aggregation.} 
To obtain the batch-wise \ConstraintLoss, 
we simply sum all the local loss terms $L(\bm{p^m}, \bm{p^n}, \bm{C^{\phi}})$  in a batch to get the total constraint loss $L_{C}$ (Eq.~\ref{eq:semanticlosssum}).
\begin{equation}
\label{eq:semanticlosssum}
L_{C} = \sum_{m}\sum_{n}\sum_{\phi}L(\bm{p^m}, \bm{p^n}, \bm{C^{\phi}})
\end{equation}

\subsection{\textit{Semantic (Sem)}}
\label{sec:semantic_loss}
In this method, we pay more attention to which specific rules in the constraint sets the pairwise local predictions should satisfy.
Our intuition is that, for each of our constraint set, good local predictions should follow one rule in that set, while bad ones may not find any rules to satisfy.
This may push the \NRE model to effectively learn from specific rules in a more focused way.

\paragraph{Representing Constraints.}
To represent the rules in the constraint sets more precisely, 
we encode each rule $\bm{c}\in\bm{C^{\phi}}$ into a single binary vector $\bm{u}$, 
thus, the whole set is represented as a vector set $\bm{U^{\phi}}$, shown as in Fig.~\ref{fig:CLC_details}. 
Again, since the rules in $\bm{C^{t*}}$ and $\bm{C^{c*}}$ have different forms, we represent them in different ways. 

For each type rule $(r_i, r_j)\in\bm{C^{t*}}$, 
the representation vector $\bm{u}$ is a binary vector whose $i^{th}$ and $j^{th}$ dimensions are set to 1 and the rest are set to 0.
Take $\bm{C^{ts}}$ in Fig.~\ref{fig:CLC_details} as an example, 
the rule $(\textit{almaMater}, \textit{city})\in\bm{C^{ts}}$ is encoded as a vector whose first two dimensions are set to 1.

For each cardinality rule $r_i\in\bm{C^{c*}}$, 
the representation vector $\bm{u}$ is a binary vector whose $i^{th}$ dimension is set to 1 and the rest are set to 0. 
In Fig.~\ref{fig:CLC_details}, the rule $\textit{almaMater}\in\bm{C^{cs}}$ is represented by a vector, where only the first dimension is set to 1. 

Different from \textit{Coherent}, here we construct one vector set to represent each sub-category constraint sets, 
resulting in 5 vector sets $\{\bm{U^{ts}}, \bm{U^{to}}, \bm{U^{tso}}, \bm{U^{cs}}, \bm{U^{co}}\}$. 
And each single rule is represented by a $|R|$-dim binary vector.

\paragraph{Local Loss Calculation.} 
Inspired by the semantic loss function (\textit{SL}) introduced in \citealp{DBLP:conf/icml/XuZFLB18}, 
which operates on a single output, we adapt the original \textit{SL} to deal with pairwise instances over different kinds of constraints. 
We design the new local loss term as:
\begin{equation}
\label{eq:semantic_loss}
L(\bm{p^m},\bm{p^n},\bm{C^{\phi}})= -I^{mn}_{\phi}log\sum_{\bm{c}\in\bm{C^{\phi}}}\mathop{f}(\bm{p^m}, \bm{p^n}, \bm{c})
\end{equation}
where $I^{mn}_{\phi}$ is an indicator same as before and $\mathop{f}$ is a score function reflecting how well the pairwise predictions match a single rule $\bm{c}\in\bm{C^{\phi}}$. 
Since the rules in $\bm{C^{t*}}$ and $\bm{C^{c*}}$ are encoded in different ways, 
we calculate $\mathop{f}$ for type constraint sets and cardinality constraint sets separately.

Thus, for a rule $\bm{c}$ in type constraint set $\bm{C^{t*}}$, the score function $f(\bm{p^m},\bm{p^n},\bm{c})$ can be calculated by:
\begin{equation}
\label{eq:sl_type}
\begin{split}
q_{i} &= p^m_i + p^n_i - p^m_i p^n_i\\
f(\bm{p^m}, \bm{p^n}, \bm{c}) &= \prod_{u_i=1}q_i\prod_{u_i=0}(1-q_i)
\end{split}
\end{equation}
where $\bm{u}$ is the vector representation of $\bm{c}$ and $q_i$ is the probability that base \NRE model predicts relation $r_i$ for at least one of  the $m^{th}$ and the $n^{th}$ instances.

For a rule $\bm{c}$ in cardinality constraint set $\bm{C^{c*}}$, $f(\bm{p^m}, \bm{p^n}, \bm{c})$ can be calculated by:
\begin{equation}
\label{eq:sl_cardinality}
f(\bm{p^m}, \bm{p^n}, \bm{c}) = \prod_{u_i=1}p^m_i p^n_i\prod_{u_i=0}(1-p^m_i p^n_i)
\end{equation}
where $p^m_i p^n_i$ means the probability that \NRE model predicts relation $r_i$ for both the $m^{th}$ and the $n^{th}$ instances.

\paragraph{Aggregation.} We use the same method as \textit{Coherent} to perform 
aggregation according to Eq.~\ref{eq:semanticlosssum}.

\bigskip
Note that \textit{Coherent} handles the constraint set as a whole and treats each single rule in that set equally, 
while \textit{Semantic} treats all rules in a constraint set as mutually exclusive and makes the pairwise predictions more satisfying one certain rule in that set.
Take $\bm{C^{cs}}$ as an example, in Eq. \ref{eq:coh_cardinality_calculation}, 
\textit{Coherent} just simply increases the probabilities of corresponding relation pairs for all positive rules, 
and each rule has the same influence on the summation. 
However, in Eq. \ref{eq:sl_cardinality}, for a potentially satisfied rule, 
\textit{Semantic} not only tries to increase the probabilities of its corresponding relation pair, 
but also lowers the probabilities of the rest. 
That is, there would not exist pair-wise local predictions which satisfy two positive rules well in one constraint set at the same time, 
since if the high probabilities of a relation pair have the positive effect on one specific rule, it has negative effect on all the others.

\section{Experiments}
\label{sec:exp}
Our experiments are designed to answer the following questions: 
(1) whether our approach can effectively utilize the relation constraints to improve the extraction performance?
(2) which \CLC module performs better, \textit{Coherent} or \textit{Semantic}?
(3) which is the better way to utilize the relation constraints, learning or post-processing?

\begin{figure*}[htbp]
	\centering
	\includegraphics[width=16.0cm]{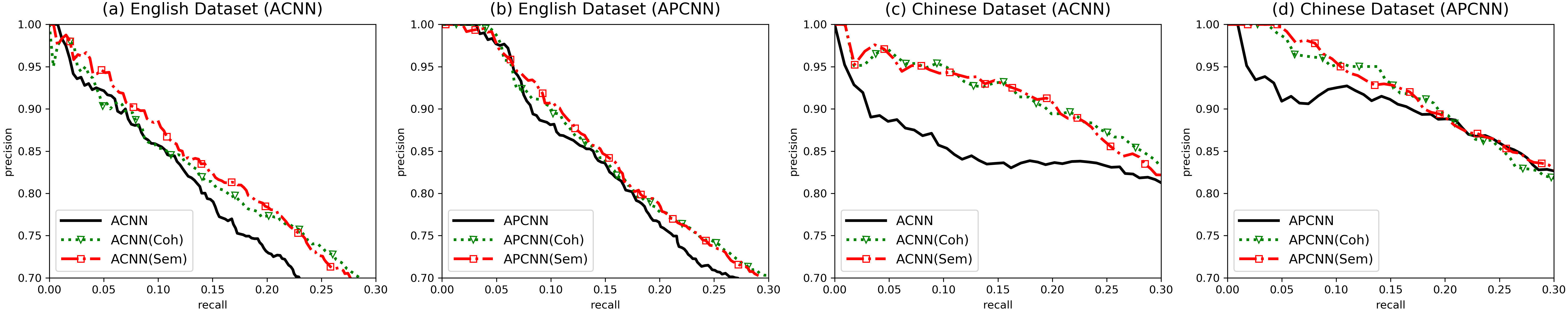}
	\caption{The PR curves of our approach on two datasets with \ACNN and \APCNN as base models.}
	\label{ex:overall_results}
\end{figure*}

\subsection{Datasets}
\label{subsec:dataset_and_evaluation_metrics}
We evaluate our approach on both English  and  Chinese datasets constructed by \citealp{chen2018encoding}.
The English one is constructed by mapping triples in DBpedia ~\cite{bizer2009dbpedia} to sentences in the New York Times Corpus. 
It has 51 relations, about 50k triples, 134k sentences for training and 30k triples, 53k sentences for testing. 
The Chinese dataset is built by mapping the triples of HudongBaiKe, a large Chinese encyclopedia, with four Chinese economic newspapers. 
It contains 28 relations, about 60k triples, 120k sentences for training and 40k triples, 83k sentences for testing.

We automatically collect relation constraints for English and Chinese datasets based on corresponding \texttt{KBs}. 
In total, we obtain 541 rules for the English dataset and 110 rules for the Chinese one. 

Here we do not use the popular \RE dataset created by~\citealp{riedel2010modeling}, since
it is produced with an earlier version of Freebase which is not available now, 
and makes it impossible to automatically collect the constraints.
Secondly, Riedel's dataset is dominated by three \textit{big} relations:
\textit{location/contains}, \textit{/people/nationality} and \textit{/people/place\_lived},
covering about 60\%  of all KB triples. 
Therefore, there are not enough data related to other relations for us to collect constraints.

\subsection{Setup} 

Following common practice in the \RE community \cite{ji2017distant,he2018see}, we report the model performance by
both precision-recall (PR) curve and Precision@N (P@N). We also report the average score of P@N (\textit{Mean}). 

The main goal of our work is to explore whether our approach can help neural models effectively learn from discrete relation constraints. 
Therefore, the first baseline models are the two most popular base \NRE models, \ACNN and \APCNN.
We also compare with the base \NRE models enhanced with a post-processing \ILP step, 
\textit{ACNN+ILP} and \textit{APCNN+ILP}, which can be considered as state-of-the-art constraint-based \RE solutions.   

We use a grid search to tune our hyper parameters, including the weight coefficient $\lambda$. 
Details about our hyper parameters are reported in Appendix.

\subsection{Main Results}
\begin{table*}[htbp]
	\centering
	\small
	\begin{tabular}{lccccc|ccccc}
		\midrule[1.5pt]
		\multicolumn{1}{c}{}& \multicolumn{5}{c}{English Dataset} & \multicolumn{5}{c}{Chinese Dataset} \\ 
		\multicolumn{1}{l}{\textbf{Model Name}}&  P@100  &   P@200   &   P@300 & \textbf{Mean}  & \multicolumn{1}{c}{$\bm{\Delta_{Base}}$} & \multicolumn{1}{c}{  P@100  } &   P@200   &   P@300   &   \textbf{Mean}  & $\bm{\Delta_{Base}}$ \\ \midrule[1.5pt]
		\textit{ACNN} & 96.70 & 92.61 & 91.72 & 93.68 & \--- & 89.08 & 86.89 & 84.52 & 86.83 & \--- \\ 
		\textit{ACNN(Coh)} & 97.39 & 93.78 & 90.69 & 93.96  & +0.3 & 95.86 & 94.86 & 93.04 & 94.59 & +7.8 \\ 
		\textit{ACNN(Sem)} & 97.62 & 95.87 & 94.12 & 95.87 & +2.2 & 95.97 & 94.61 & 93.53 & 94.70 & +8.1 \\ 
		\textit{ACNN+ILP} & 97.87 & 94.36 & 93.16 & 95.13 & +1.5 & 93.75 & 92.18 & 90.10 & 92.01 & +5.2\\
 		\textit{ACNN(Coh)+ILP} & 97.73 & 94.51 & 91.29 & 94.51 & +0.8 & 97.09 & 96.18 & 94.01 & 95.76 & +9.0 \\ 
 		\textit{ACNN(Sem)+ILP} & \textbf{98.17} & \textbf{96.6} & \textbf{95.48} & \textbf{96.75} & \textbf{+3.1} &  \textbf{97.73} & \textbf{96.40} & \textbf{94.43} & \textbf{96.18} & \textbf{+9.4} \\
		\midrule[1pt]
		\textit{APCNN} & 100 & 98.97 & 97.41 & 98.79 & \--- & 92.96 & 91.75 & 91.08 & 91.93 & \--- \\ 
		\textit{APCNN(Coh)} & 100 & 99.57 & 97.33 & 98.97 & +0.2 &98.88 & 96.00 & 94.98 & 96.62 & +4.7 \\ 
		\textit{APCNN(Sem)} & 100 & 100 & 97.95 & 99.32 & +0.5 & 100 & 96.97 & 93.42 & 96.80 & +4.9 \\
		\textit{APCNN+ILP} & 100 & 99.13 & 97.55 & 98.89 & +0.1 & 96.06 & 95.15 & 94.63 & 95.28 & +3.4 \\ 
 		\textit{APCNN(Coh)+ILP} & 100 & 100 & 98.03 & 99.34 & +0.6 & 99.07 & 96.17 & \textbf{95.16} & 96.79 & +4.9 \\ 
 		\textit{APCNN(Sem)+ILP} & \textbf{100} & \textbf{100} & \textbf{98.39} & \textbf{99.46}& \textbf{+0.7} & \textbf{100} & \textbf{97.67} & 94.25 & \textbf{97.31} & \textbf{+5.4} \\
		\midrule[1.5pt]
	\end{tabular}
	\caption{Summary P@N(\%) scores of our approach on two datasets with \ACNN and \APCNN as base models. 
		$\bm{\Delta_{Base}}$ indicates the difference between mentioned model and the base \NRE model (\ACNN in the top and \APCNN in the bottom). And the name with \textit{+ILP} means that we perform \ILP over the model's outputs as an extra post-processing.}
	\label{Table:main_results}
\end{table*}

Our main results are summarized in Fig.~\ref{ex:overall_results} and Table~\ref{Table:main_results}. 
As shown in Fig.~\ref{ex:overall_results}, we can see that both the red and green dot lines are lifted above the solid black lines, 
showing that after equipped with our \CLC modules, i.e., \textit{Coherent} and \textit{Semantic}, 
both \ACNN and \APCNN obtain significant improvement on the English and Chinese datasets. 
This indicates our \CLC module actually helps the base \NRE models benefit from properly utilizing the relation constraints, without interference to the base models.

However, we find that our approach obtains different levels of improvement on the two datasets.  
On the Chinese one, as shown in Table~\ref{Table:main_results}, with our \textit{Semantic} version \CLC,  
\textit{APCNN(Sem)} gains 4.9\% improvement in \textit{Mean} compared to \textit{APCNN}, 
but, on the English dataset, it only receives 0.5\%  in \textit{Mean}. 
Similar trends are also found for the \textit{Coherent} version and the \textit{ACNN} base model.
The better performance gain on the Chinese dataset is mainly
because its relation definitions are more clear compared to that of the English dataset. 
For example, in English dataset, there are 8 relations whose object could be any \textsc{location}, 
such as \textit{birthPlace}, while only 3 similar relations exist in Chinese dataset.

In addition, we investigate the performance improvement when applying our \CLC module to different base \NRE models. 
Although both \ACNN and \APCNN are improved by our \CLC module in various datasets, 
we can still observe that \ACNN generally receives more performance improvement compared with the \APCNN base model.  
Taking the \textit{Semantic} method as an example, as shown in Table~\ref{Table:main_results}, 
on the English dataset, \textit{ACNN(Sem)} obtains 2.2\% performance improvement in \textit{Mean} against \ACNN, 
while \textit{APCNN(Sem)} only fetches 0.5\% improvement. 
And similar trends can be found in the \textit{Coherent} method and on the Chinese dataset.
The more improvement when taking the \ACNN as base \NRE model is because, compared with \ACNN, 
\APCNN itself is designed to take the entity-aware sentence structure information into account, 
thus can extract more effective features that, to some extent, can implicitly capture part of the arguments' type and cardinality requirements of a relation, 
leaving relatively less space for our \CLC module to improve. 

\subsubsection{Comparing \textit{Coherent} and \textit{Semantic}}
This paper presents two different methods, \textit{Coh} and \textit{Sem},  
to represent and integrate the relation constraints, 
both of which can lead to substantial improvement with both base models and datasets.  
Specifically, as shown in Table~\ref{Table:main_results}, 
\textit{Sem} brings slightly more improvement than \textit{Coh} in most of the settings, 
e.g., on Chinese dataset, \textit{APCNN(Sem)} obtain about 0.2\% more gains (4.9\% vs 4.7\%) in \textit{Mean} than \textit{APCNN(Coh)}. 
We think the reason is that \textit{Sem} provides a more precise treatment for the constraints, 
e.g., embedding each rule with a vector and trying to evaluate the \NRE output against one specific rule, 
while \textit{Coh} represents all rules in a sub-category with one single vector and evaluates the output against whole set of rules,  
which is admittedly a more coarse fashion. 

\subsubsection{Learning? or Post-processing?}
Previous works show that \ILP can effectively solve the inconsistency among predictions in a post-processing fashion \cite{chen2018encoding}. 

Now we discuss which is the better way to utilize the relation constraints, 
our \CLC module or traditional \ILP post-processing.  
As shown in Table~\ref{Table:main_results}, 
both  \textit{APCNN(Sem)} and \textit{APCNN(Coh)} outperforms \textit{APCNN+ILP} by at least 0.1\% on the English dataset and 1.0\% on the Chinese dataset. 
Similar trends can be also found for \textit{ACNN(Sem)} and \textit{ACNN(Coh)}.
This shows that helping base \NRE models to learn from the relation constraints can generally bring more improvement, 
thus utilizes the  constraints more effectively compared to utilizing those constraints in a post-processing way. 

We can also apply \ILP as a post-processing step to our approach,
since our \CLC module works in the model training phase, and leaves the testing phase as it is. 
Interestingly, as shown in Table~\ref{Table:main_results}, with an extra \ILP post-processing, 
both \textit{Coh} and \textit{Sem} obtain further improvement with different base \NRE models on different datasets. 
This indicates that our \CLC module still may not fully exploit the useful information behind the relation constraints. 
The reasons may be that our approach and the \ILP post-processing exploit the relation constraints from different perspectives. 
For example, our \CLC operates in a mini-batch level during training,  that is a relatively local view, 
but \ILP post-processing directly optimizes the model output in a slightly global view.    

Moreover, in Table \ref{Table:mean_delta},
we find that applying \ILP to our \CLC enhanced model receives relatively less gain compared to applying \ILP to the base model, 
e.g., 0.5\% for \textit{APCNN(Sem)} v.s. 3.4\% for \APCNN on the Chinese dataset.  
This observation may indicate that our approach has pushed \NRE base models to learn part of the useful information behind relation constraints, 
leaving fewer inconsistent outputs for \ILP post-processing to filter out. 
On the other hand, this observation shows again that  our \CLC approach and the \ILP post-processing exploit complementary aspects from the relation constraints, 
and our \CLC module could be further improved by taking more global optimization into account. 

\begin{table}[htbp]
	\centering
	\small
	\begin{tabular}{lcccc}
		\midrule[1.5pt]
		& \multicolumn{2}{c}{English} & \multicolumn{2}{c}{Chinese} \\
		& Mean & $\bm{\Delta_{ILP}}$ & Mean & $\bm{\Delta_{ILP}}$ \\
		\hline
		\textit{ACNN} & 93.68 & +1.5 & 86.83 & +5.2 \\
		\textit{ACNN(Coh)} & 93.86 & +0.6 & 94.59 & +1.2 \\
		\textit{ACNN(Sem)} & \textbf{95.87} & +0.9 & \textbf{94.70} & +1.5 \\
		\hline
		\textit{APCNN} & 98.79 & +0.1 & 91.93 & +3.4 \\
		\textit{APCNN(Coh)} & 98.97 & +0.4 & 96.62 & +0.2 \\
		\textit{APCNN(Sem)} & \textbf{99.32} & +0.1 & \textbf{96.80} & +0.5 \\
		\bottomrule[1.5pt]
	\end{tabular}
	\caption{Relative improvement of different models in \textit{Mean}. $\bm{\Delta_{ILP}}$ is the performance difference between the mentioned model and the same model with an extra \ILP step. For example, $\bm{\Delta_{ILP}}$ corresponding to raw \textit{ACNN} indicates that applying \ILP to \textit{ACNN} obtains 1.5\% and 5.2\% gain in \textit{Mean} on English and Chinese dataset, respectively.}
	\label{Table:mean_delta}
\end{table}

\subsection{More Analysis}
To better understand what our approach learns from the constraints, we  
take a deep look at the  outputs of \textit{APCNN} and \textit{APCNN(Sem)} on the test split of the Chinese dataset. 
First, we count the total number of contradictory pairwise predictions
and find that applying our \textit{Semantic} method to \textit{APCNN} achieves a reduction of 5,966 violations, 28.0\% of the total\footnote{Detailed numbers per category are reported in Appendix.}.
This indicates our approach has pushed the base \NRE models to learn from the relation constraints. However, 
there are still many remaining violations since our approach operates during training in a soft and local way, compared to \ILP during testing.  

Another observation is that our approach actually reduces the violations related to each relations, 
and especially does better when there are tighter requirements on the relation's arguments. 
For example, \textit{APCNN(Sem)} reduces 89.6\% violations for relation \textit{locationState} compared to \textit{APCNN}, 
but for \textit{locationRegion}, it only reduces 36.3\%.
This is because the relation constraints may indicate more clear  arguments' type requirements for \textit{locationState} than those of \textit{locationRegion},
which are captured by our \CLC module to push into the base \NRE during training.

 \section{Conclusion}
\label{sec:conclusion}
In this paper, we propose a unified framework to effectively integrate discrete relation constraints with neural networks for relation extraction. 
Specifically, we develop two approaches to evaluate how well \NRE predictions satisfy our relation constraints in a batch-wise, 
from both general and precise perspectives. 
We explore our approach on English and Chinese dataset,
and the experimental results show that our approach can help the base \NRE models to effectively learn from the discrete relation constraints, 
and outperform popular \NRE models as well as their \ILP enhanced versions. 
Our study reveals that learning with the constraints can better utilize the constraints from a different perspective compared to the \ILP post-processing method.   

\section{Acknowledgments}
We thank anonymous reviewers for their valuable suggestions. This work is supported in part by the National Hi-Tech R\&D Program of China (2018YFC0831900) and the NSFC Grants (No.61672057, 61672058). For any correspondence, please contact Yansong Feng.

\appendix
\section{Appendix}
\subsection{Parameter Settings}
In the experiment, both \ACNN and \APCNN use convolution window size 3, sentence embedding size 256, position embedding size 5 and batch size 50.
The word embedding size is 50 and 300 for the English and Chinese dataset, respectively. We use Adam with learning rate 0.001 to train our model.
And we fine-tune the constraint loss coefficients for each experimental settings, reported in Table \ref{Table:1}
\begin{table}[htbp]
    \small
    \centering
    \begin{tabular}{lcccc}
        \midrule[1.5pt]
        \multicolumn{1}{c}{} & \multicolumn{2}{c}{\textbf{English dataset}} & \multicolumn{2}{c}{\textbf{Chinese dataset}} \\ 
         & \textbf{\textit{ACNN}} & \textbf{\textit{APCNN}} & \textbf{\textit{ACNN}} & \textbf{\textit{APCNN}}    \\ \midrule[0.75pt]
        \textbf{\textit{Sem}}  & $1 \times 10^{-3}$ & $1 \times 10^{-4}$ & $5 \times 10^{-4}$ & $1 \times 10^{-5}$  \\ 
        \textbf{\textit{Coh}} & $5 \times 10^{-3}$ & $1 \times 10^{-4}$  & $1 \times 10^{-4}$  & $1 \times 10^{-4}$  \\
        \midrule[1.5pt]
    \end{tabular}
    \caption{The value of coeffecient $\lambda$ for each experimental settings.}
    \label{Table:1}
\end{table}

\subsection{Local Loss Calculating Indicators}

In this section, we list the assignment methods for all $I^{mn}_{\phi}$ which indicates whether to calculate local loss term $L(\bm{p^m}, \bm{p^n}, \bm{C^{\phi}})$ for the combination of the $m^{th}$ and $n^{th}$ instances, $<subj_m, r_m, obj_m>$ and $<subj_n, r_n, obj_n>$, within a batch.\\
    \begin{equation}
        I^{mn}_{ts}=\left\{
        \begin{aligned}
        1 & , & subj_m=subj_n; \\
        0 & , & subj_m \ne subj_n. 
        \end{aligned}
        \right.
    \end{equation}
where $subj_m=subj_n$ means the two triples have the same subject type, thus, corresponding predicted relation pair may be contradictory with $\bm{C^{ts}}$.
    \begin{equation}
        I^{mn}_{to}=\left\{
        \begin{aligned}
        1 & , & obj_m=obj_n; \\
        0 & , & obj_m \ne obj_n. 
        \end{aligned}
        \right.
    \end{equation}
where $obj_m=obj_n$ means the two triples have the same object type, thus, corresponding predicted relation pair may be contradictory with $\bm{C^{to}}$.
    \begin{equation}
        I^{mn}_{tso}=\left\{
        \begin{aligned}
        1 & , & subj_m=obj_n || obj_m=subj_n; \\
        0 & , & otherwise.
        \end{aligned}
        \right.
    \end{equation}
where $subj_m=obj_n || obj_m=subj_n$ means the subject type of one relation is same as the object type of the other, thus, corresponding predicted relation pair may be contradictory with $\bm{C^{tso}}$.
    \begin{equation}
        I^{mn}_{cs}=\left\{
        \begin{aligned}
        1 & , & subj_m \ne subj_n \&\& obj_m=obj_n; \\
        0 & , & otherwise.
        \end{aligned}
        \right.
    \end{equation}
where $subj_m \ne subj_n \&\& obj_m=obj_n$ means that for a given object, there are multiple subjects, thus, corresponding predicted relation pair may be contradictory with $\bm{C^{cs}}$.
    \begin{equation}
        I^{mn}_{co}=\left\{
        \begin{aligned}
        0 & , & obj_m \ne obj_n \&\& subj_m=subj_n; \\
        1 & , & otherwise.
        \end{aligned}
        \right.
    \end{equation}
where $obj_m \ne obj_n \&\& subj_m=subj_n$ means that for a given subject, there are multiple objects, thus, corresponding predicted relation pair may be contradictory with $\bm{C^{co}}$.

\subsection{Statistics on Violations for Each Constraint Set}
In this section, we collect the number of violations for each specific constraint set among the relation predictions of \textit{APCNN} and \textit{APCNN(Sem)} on Chinese dataset, shown as in Table \ref{Table:3}.

\begin{table}[htbp]
    \centering
    \tiny
    \begin{tabular}{lcccccc}
        \toprule[1.5pt]
        & $\bm{C^{ts}}$ & $\bm{C^{to}}$ & $\bm{C^{tso}}$ & $\bm{C^{cs}}$ & $\bm{C^{co}}$ & \textbf{Total} \\ \midrule[0.75pt]
        \textit{APCNN} & 850 & 11,183 & 7,636 & 1,464 & 209 & 21,342 \\ 
        \textit{APCNN(Sem)} & 596 & 6,772  & 6,573 & 1,259 & 176 & 15,376 \\ 
        \midrule[0.75pt]
    \end{tabular}
    \caption{Statistics on predicted relation pairs which are contradictory with constraint set $\bm{C^{\phi}}$ for test data of Chinese dataset. 
    }
    \label{Table:3}
\end{table}

\subsection{Further Discussions on Training Procedure}
First, adjusting the coefficient $\lambda$ of our constraint loss by a dynamic mechanism during training would be helpful.
Particularly, we use Eq.\ref{Eq:1} to dynamic adjust $\lambda$.
\begin{equation}
    \lambda = -\alpha * \frac{2*|E_{cur} - 0.5*E_{total}|}{E_{total}} + \alpha
    \label{Eq:1}
\end{equation}
where $\alpha$ is a constant which represents the max value of $\lambda$, $E_{cur}$ and $E_{total}$ represent the current epoch number and the total epoch number, respectively.
By Eq.\ref{Eq:1}, we make $\lambda$ first rise and then fall, since we think the \NRE model should more focus on the original loss at the start of training, and the influence of relation constraints should decrease after the \NRE model has learned relation constraints pretty well. 
We apply this dynamic mechanism on English dataset with \APCNN as base model, and achieve 99.33\% compared to 99.32\% of constant $\lambda$ in \textit{Mean}. We think may be a more nicely dynamic mechanism which captures the inherent of combining relation constraints with \NRE models could fetch more improvement.

In addition, organizing related instances into a same mini-batch seems to be helpful too, while how to make the reorganized data evenly distributed and maintaining the randomness of data at the same time, is very challenging. We leave this modification into the future work.

\bibliography{YeYCL}

\begin{thebibliography}{}

\bibitem[\protect\citeauthoryear{Bizer \bgroup et al\mbox.\egroup
  }{2009}]{bizer2009dbpedia}
Bizer, C.; Lehmann, J.; Kobilarov, G.; Auer, S.; Becker, C.; Cyganiak, R.; and
  Hellmann, S.
\newblock 2009.
\newblock Dbpedia-a crystallization point for the web of data.
\newblock {\em Web Semantics: science, services and agents on the world wide
  web} 7(3):154--165.

\bibitem[\protect\citeauthoryear{Chen \bgroup et al\mbox.\egroup
  }{2018}]{chen2018encoding}
Chen, L.; Feng, Y.; Huang, S.; Luo, B.; and Zhao, D.
\newblock 2018.
\newblock Encoding implicit relation requirements for relation extraction: A
  joint inference approach.
\newblock {\em Artificial Intelligence} 265:45--66.

\bibitem[\protect\citeauthoryear{Dai, Li, and Xu}{2016}]{dai2016cfo}
Dai, Z.; Li, L.; and Xu, W.
\newblock 2016.
\newblock Cfo: Conditional focused neural question answering with large-scale
  knowledge bases.
\newblock In {\em Proceedings of the 54th Annual Meeting of the Association for
  Computational Linguistics (Volume 1: Long Papers)}, volume~1,  800--810.

\bibitem[\protect\citeauthoryear{Feng \bgroup et al\mbox.\egroup
  }{2018}]{feng2018reinforcement}
Feng, J.; Huang, M.; Zhao, L.; Yang, Y.; and Zhu, X.
\newblock 2018.
\newblock Reinforcement learning for relation classification from noisy data.
\newblock In {\em Thirty-Second AAAI Conference on Artificial Intelligence}.

\bibitem[\protect\citeauthoryear{He \bgroup et al\mbox.\egroup
  }{2018}]{he2018see}
He, Z.; Chen, W.; Li, Z.; Zhang, M.; Zhang, W.; and Zhang, M.
\newblock 2018.
\newblock See: Syntax-aware entity embedding for neural relation extraction.
\newblock In {\em Thirty-Second AAAI Conference on Artificial Intelligence}.

\bibitem[\protect\citeauthoryear{Hoffmann \bgroup et al\mbox.\egroup
  }{2011}]{hoffmann2011knowledge}
Hoffmann, R.; Zhang, C.; Ling, X.; Zettlemoyer, L.; and Weld, D.~S.
\newblock 2011.
\newblock Knowledge-based weak supervision for information extraction of
  overlapping relations.
\newblock In {\em Proceedings of ACL},  541--550.

\bibitem[\protect\citeauthoryear{Hu \bgroup et al\mbox.\egroup
  }{2016}]{hu2016harnessing}
Hu, Z.; Ma, X.; Liu, Z.; Hovy, E.; and Xing, E.
\newblock 2016.
\newblock Harnessing deep neural networks with logic rules.
\newblock {\em arXiv preprint arXiv:1603.06318}.

\bibitem[\protect\citeauthoryear{Ji \bgroup et al\mbox.\egroup
  }{2017}]{ji2017distant}
Ji, G.; Liu, K.; He, S.; and Zhao, J.
\newblock 2017.
\newblock Distant supervision for relation extraction with sentence-level
  attention and entity descriptions.
\newblock In {\em Thirty-First AAAI Conference on Artificial Intelligence}.

\bibitem[\protect\citeauthoryear{Jia \bgroup et al\mbox.\egroup
  }{2019}]{jia2019arnor}
Jia, W.; Dai, D.; Xiao, X.; and Wu, H.
\newblock 2019.
\newblock Arnor: Attention regularization based noise reduction for distant
  supervision relation classification.
\newblock In {\em Proceedings of the 57th Conference of the Association for
  Computational Linguistics},  1399--1408.

\bibitem[\protect\citeauthoryear{Lai \bgroup et al\mbox.\egroup
  }{2019}]{lai2019lattice}
Lai, Y.; Feng, Y.; Yu, X.; Wang, Z.; Xu, K.; and Zhao, D.
\newblock 2019.
\newblock Lattice cnns for matching based chinese question answering.
\newblock {\em arXiv preprint arXiv:1902.09087}.

\bibitem[\protect\citeauthoryear{Lin \bgroup et al\mbox.\egroup
  }{2016}]{lin2016neural}
Lin, Y.; Shen, S.; Liu, Z.; Luan, H.; and Sun, M.
\newblock 2016.
\newblock Neural relation extraction with selective attention over instances.
\newblock In {\em Proceedings of the 54th Annual Meeting of the Association for
  Computational Linguistics (Volume 1: Long Papers)}, volume~1,  2124--2133.

\bibitem[\protect\citeauthoryear{Luo \bgroup et al\mbox.\egroup
  }{2018}]{luo2018marrying}
Luo, B.; Feng, Y.; Wang, Z.; Huang, S.; Yan, R.; and Zhao, D.
\newblock 2018.
\newblock Marrying up regular expressions with neural networks: A case study
  for spoken language understanding.
\newblock In {\em Proceedings of the 56th Annual Meeting of the Association for
  Computational Linguistics (Volume 1: Long Papers)}, volume~1,  2083--2093.

\bibitem[\protect\citeauthoryear{Mintz \bgroup et al\mbox.\egroup
  }{2009}]{mintz2009distant}
Mintz, M.; Bills, S.; Snow, R.; and Jurafsky, D.
\newblock 2009.
\newblock Distant supervision for relation extraction without labeled data.
\newblock In {\em Proceedings of the Joint Conference of the 47th Annual
  Meeting of the ACL and the 4th International Joint Conference on Natural
  Language Processing of the AFNLP: Volume 2-Volume 2},  1003--1011.
\newblock Association for Computational Linguistics.

\bibitem[\protect\citeauthoryear{Qin, Xu, and Wang}{2018}]{qin2018dsgan}
Qin, P.; Xu, W.; and Wang, W.~Y.
\newblock 2018.
\newblock Dsgan: generative adversarial training for distant supervision
  relation extraction.
\newblock {\em arXiv preprint arXiv:1805.09929}.

\bibitem[\protect\citeauthoryear{Riedel, Yao, and
  McCallum}{2010}]{riedel2010modeling}
Riedel, S.; Yao, L.; and McCallum, A.
\newblock 2010.
\newblock Modeling relations and their mentions without labeled text.
\newblock In {\em Joint European Conference on Machine Learning and Knowledge
  Discovery in Databases},  148--163.
\newblock Springer.

\bibitem[\protect\citeauthoryear{Suchanek \bgroup et al\mbox.\egroup
  }{2013}]{suchanek2013advances}
Suchanek, F.; Fan, J.; Hoffmann, R.; Riedel, S.; and Talukdar, P.~P.
\newblock 2013.
\newblock Advances in automated knowledge base construction.
\newblock {\em SIGMOD Records journal, March}.

\bibitem[\protect\citeauthoryear{Surdeanu \bgroup et al\mbox.\egroup
  }{2012}]{surdeanu2012multi}
Surdeanu, M.; Tibshirani, J.; Nallapati, R.; and Manning, C.~D.
\newblock 2012.
\newblock Multi-instance multi-label learning for relation extraction.
\newblock In {\em Proceedings of the 2012 joint conference on empirical methods
  in natural language processing and computational natural language learning},
  455--465.
\newblock Association for Computational Linguistics.

\bibitem[\protect\citeauthoryear{Vashishth \bgroup et al\mbox.\egroup
  }{2018}]{vashishth2018reside}
Vashishth, S.; Joshi, R.; Prayaga, S.~S.; Bhattacharyya, C.; and Talukdar, P.
\newblock 2018.
\newblock Reside: Improving distantly-supervised neural relation extraction
  using side information.
\newblock {\em arXiv preprint arXiv:1812.04361}.

\bibitem[\protect\citeauthoryear{Wu, He, and Hu}{2018}]{wu2018entity}
Wu, G.; He, Y.; and Hu, X.
\newblock 2018.
\newblock Entity linking: an issue to extract corresponding entity with
  knowledge base.
\newblock {\em IEEE Access} 6:6220--6231.

\bibitem[\protect\citeauthoryear{Xu \bgroup et al\mbox.\egroup
  }{2016}]{xu2016acl}
Xu, K.; Reddy, S.; Feng, Y.; Huang, S.; and Zhao, D.
\newblock 2016.
\newblock Question answering on {F}reebase via relation extraction and textual
  evidence.
\newblock In {\em Proceedings of the 54th Annual Meeting of the Association for
  Computational Linguistics (Volume 1: Long Papers)},  2326--2336.
\newblock Berlin, Germany: Association for Computational Linguistics.

\bibitem[\protect\citeauthoryear{Xu \bgroup et al\mbox.\egroup
  }{2018}]{DBLP:conf/icml/XuZFLB18}
Xu, J.; Zhang, Z.; Friedman, T.; Liang, Y.; and den Broeck, G.~V.
\newblock 2018.
\newblock A semantic loss function for deep learning with symbolic knowledge.
\newblock In {\em Proceedings of the 35th International Conference on Machine
  Learning, {ICML} 2018, Stockholmsm{\"{a}}ssan, Stockholm, Sweden, July 10-15,
  2018},  5498--5507.

\bibitem[\protect\citeauthoryear{Yu \bgroup et al\mbox.\egroup
  }{2017}]{yu2017acl}
Yu, M.; Yin, W.; Hasan, K.~S.; dos Santos, C.; Xiang, B.; and Zhou, B.
\newblock 2017.
\newblock Improved neural relation detection for knowledge base question
  answering.
\newblock In {\em Proceedings of the 55th Annual Meeting of the Association for
  Computational Linguistics (Volume 1: Long Papers)},  571--581.
\newblock Vancouver, Canada: Association for Computational Linguistics.

\bibitem[\protect\citeauthoryear{Zeng \bgroup et al\mbox.\egroup
  }{2014}]{zeng2014relation}
Zeng, D.; Liu, K.; Lai, S.; Zhou, G.; and Zhao, J.
\newblock 2014.
\newblock Relation classification via convolutional deep neural network.
\newblock In {\em Proceedings of COLING 2014, the 25th International Conference
  on Computational Linguistics: Technical Papers},  2335--2344.

\bibitem[\protect\citeauthoryear{Zeng \bgroup et al\mbox.\egroup
  }{2015}]{zeng2015distant}
Zeng, D.; Liu, K.; Chen, Y.; and Zhao, J.
\newblock 2015.
\newblock Distant supervision for relation extraction via piecewise
  convolutional neural networks.
\newblock In {\em EMNLP},  1753--1762.

\end{thebibliography}
\bibliographystyle{aaai}
\end{document}